\documentclass[letterpaper]{article} % DO NOT CHANGE THIS
\usepackage{aaai2026}  % DO NOT CHANGE THIS
\usepackage{times}  % DO NOT CHANGE THIS
\usepackage{helvet}  % DO NOT CHANGE THIS
\usepackage{courier}  % DO NOT CHANGE THIS
\usepackage[hyphens]{url}  % DO NOT CHANGE THIS
\usepackage{graphicx} % DO NOT CHANGE THIS
\usepackage{natbib}  % DO NOT CHANGE THIS AND DO NOT ADD ANY OPTIONS TO IT
\usepackage{caption} % DO NOT CHANGE THIS AND DO NOT ADD ANY OPTIONS TO IT
\usepackage{algorithm}
\usepackage{algorithmic}
\usepackage{pifont}
\usepackage{amsmath}
\usepackage{amssymb}
\usepackage{fdsymbol}
\usepackage{placeins}
\usepackage{tikz}
\usepackage{colortbl} 
\usepackage{nicefrac}
\usepackage{dsfont}
\usepackage{marvosym}
\usepackage{multirow}
\usepackage{nicematrix}
\usepackage{booktabs}
\usepackage{pifont}
\newcommand{\best}[1]{\textbf{\textcolor{red}{#1}}}
\newcommand{\secondbest}[1]{\underline{\textcolor{blue}{#1}}}

\newcommand\shline{\specialrule{0.8pt}{0pt}{0pt}}

\definecolor{OursColor}{HTML}{FFFFCC}  % Lighter yellow
\definecolor{myIDBcolor}{HTML}{FFF5F0}
\definecolor{myCCDBcolor}{HTML}{F5FFF0}  % Lighter green

\usepackage{newfloat}
\usepackage{listings}
\usepackage{graphicx}
\DeclareCaptionStyle{ruled}{labelfont=normalfont,labelsep=colon,strut=off} % DO NOT CHANGE THIS
\floatstyle{ruled}
\newfloat{listing}{tb}{lst}{}
\floatname{listing}{Listing}
\nocopyright
\title{VQ-Insight: Teaching VLMs for AI-Generated Video Quality Understanding via Progressive Visual Reinforcement Learning}
\author{
    Xuanyu Zhang\textsuperscript{\rm 1,\rm 2,}\equalcontrib,
    Weiqi Li\textsuperscript{\rm 1,\rm 2,}\equalcontrib,
    Shijie Zhao\textsuperscript{\rm 2,$\vardiamondsuit$,\Letter},
    Junlin Li\textsuperscript{\rm 2},
    Li Zhang\textsuperscript{\rm 2},
    Jian Zhang\textsuperscript{\rm 1,\Letter}
}
\affiliations{
    \textsuperscript{\rm 1}School of Electronic and Computer Engineering, Peking University,  
    \textsuperscript{\rm 2}ByteDance Inc.\\
}

\usepackage{bibentry}
\begin{document}

\maketitle
\renewcommand*{\thefootnote}{$\vardiamondsuit$}
\footnotetext[1]{Project Lead. \Letter: Corresponding authors.}

\begin{abstract}
Recent advances in AI-generated content (AIGC) have led to the emergence of powerful text-to-video generation models. Despite these successes, evaluating the quality of AIGC-generated videos remains challenging due to limited generalization, lack of temporal awareness, heavy reliance on large-scale annotated datasets, and the lack of effective interaction with generation models. Most current approaches rely on supervised finetuning of vision-language models (VLMs), which often require large-scale annotated datasets and tend to decouple understanding and generation. To address these shortcomings, we propose VQ-Insight, a novel reasoning-style VLM framework for AIGC video quality assessment. Our approach features: (1) a progressive video quality learning scheme that combines image quality warm-up, general task-specific temporal learning, and joint optimization with the video generation model; (2) the design of multi-dimension scoring rewards, preference comparison rewards, and temporal modeling rewards to enhance both generalization and specialization in video quality evaluation. Extensive experiments demonstrate that VQ-Insight consistently outperforms state-of-the-art baselines in preference comparison, multi-dimension scoring, and natural video scoring, bringing significant improvements for video generation tasks.
\end{abstract}

\section{Introduction}

In recent years, AI-generated content (AIGC) has demonstrated remarkable progress in video generation, giving rise to a variety of powerful text-to-video generative models~\cite{hongcogvideo, yang2024cogvideox, zheng2024open, chen2024videocrafter2,li2024omnidrag}, such as Sora, Runway Gen-2, and Pika. These models have shown significant potential in producing longer-duration videos with higher quality and improved naturalness. Despite substantial advances, generated videos from these models still frequently suffer from issues including unnaturalness, consistency errors, and poor alignment with human preferences, significantly hindering their practical application. Consequently, establishing reliable evaluation approaches for AIGC-generated videos is of crucial importance. Such evaluation methodologies can not only facilitate fine-grained manipultation of generated contents, but also serve as a robust basis for Reinforcement Learning with Human Feedback (RLHF), guiding models to more closely match user expectations. 

\begin{figure*}[t!]
	\centering    \includegraphics[width=1.0\linewidth]{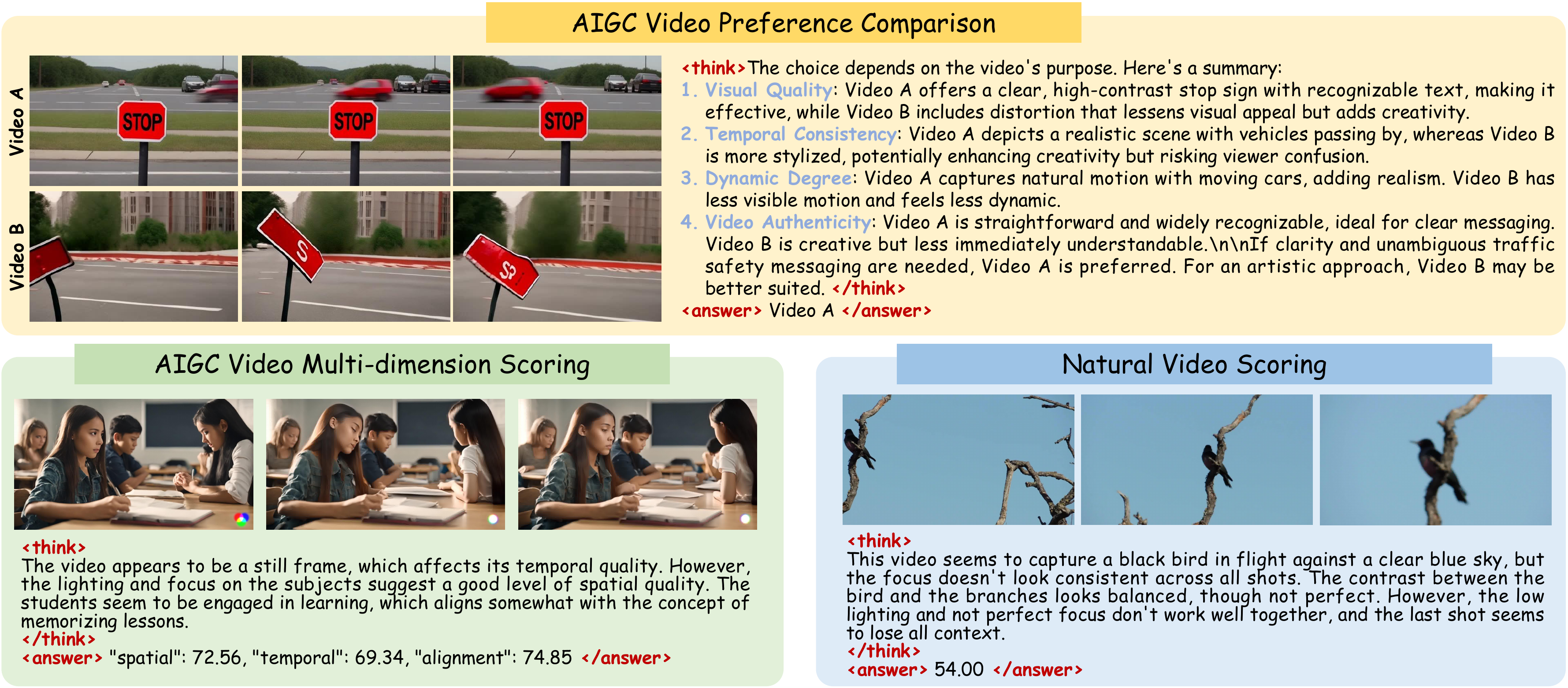}
	\vspace{-15pt}
	\caption{We propose a reasoning-style vision-language model \textbf{VQ-Insight}, which accurately performs AIGC video preference comparison, AIGC video multi-dimension scoring, and natural video scoring, accompanied by detailed and reasonable reasoning processes. Our VQ-Insight can be applied to post-training of video generation models and zero-shot content repairing.}
	\label{teasor}
\vspace{-10pt}
\end{figure*}

% A key bottleneck in applying RLHF to video generation tasks lies in the design of effective reward models. Specifically, reward models for visual tasks can be categorized into two paradigms: pointwise scoring, which assigns a continuous numerical score to a single video and supports gradient-based optimization frameworks like GRPO~\cite{guo2025deepseek} and PPO~\cite{schulman2017proximal}; pairwise ranking, which compares two videos to generate preference data for RL methods like DPO~\cite{rafailov2023direct}. Handcrafted or heuristic-based reward models often struggle to effectively balance these two requirements, and fail to accurately capture essential factors governing AIGC video generation, thus providing limited or inaccurate signals for model improvement. Recently, with advancements of vision-large language models (VLMs), researchers have begun to leverage their enhanced understanding of multimodal content and human preferences for video assessment, resulting in significant improvements. For instance, VideoScore~\cite{he2024videoscore} firstly finetuned a VLM on VideoFeedback to simulate human preference, enabling the model to perform both scoring and preference comparisons. Furthermore, VisionReward~\cite{xu2024visionreward} converted multi-dimension visual question-answering results into an interpretable binary format and employed a linear regression head to predict preference scores. However, all these methods fail to effectively balance accuracy, generalizability, and interpretability in AIGC video evaluation.

A key challenge in applying RLHF to video generation lies in the design of effective video evaluation models. Existing alignment methods for video generation typically either assign a continuous numerical score to a single video, supporting gradient-based optimization frameworks such as GRPO~\cite{guo2025deepseek,liu2025flow, jiang2025t2i, you2025teaching}, or compare two videos to generate preference data for reinforcement learning methods like DPO~\cite{rafailov2023direct, liu2025videodpo}. While significant progress has been made in evaluating natural images~\cite{li2025q, you2025teaching} and videos~\cite{jia2024vqa}, the assessment of AIGC videos remains largely underexplored. Compared to natural video quality assessment, AIGC video evaluation presents unique challenges. \textbf{1):} Due to the inherent instability of generative models and diversity of generation requirements, AIGC videos often require more fine-grained evaluation criteria; \textbf{2):} The rapid evolution of generation techniques calls for an evaluation mechanism that can quickly adapt to biases from different models and annotators. Recent works have begun leveraging vision-language models (VLMs) to address this gap. For instance, VideoScore~\cite{he2024videoscore} finetunes VLMs to support both scoring and ranking, while VisionReward~\cite{xu2024visionreward} converts multi-dimensional visual question answering outputs into preference signals. However, existing approaches still face limitations in balancing accuracy, generalizability, and interpretability in AIGC video evaluation.

Existing VLM-based AIGC video evaluation methods primarily rely on supervised finetuning (SFT), forcefully training the large models to regress video quality scores or directly judge human preferences. This approach suffers from three main drawbacks. \textbf{First}, it significantly diminishes the visual perception and general reasoning abilities of a general agent, reducing it to merely a scoring specialist. However, since different human annotators often exhibit biases when scoring the same video, simply regressing scores can in some sense be meaningless. We would prefer to inspire the model's intrinsic potential for better understanding of AIGC video quality by teaching it scoring and preference comparison tasks. \textbf{Second}, existing methods~\cite{wang2025unified, wang2025unifiedcot} typically require massive amounts of training data and continual construction of new benchmarks to keep pace with the rapidly evolving AIGC video generation methods. For instance, VisionReward~\cite{xu2024visionreward} employed 80k visual question-answering annotations along with 2k preference comparisons to simulate human preferences, and VideoAlign~\cite{liu2025improving} even constructed a 182k human-labeled preference training samples, consuming enormous human and material resources. However, despite the variety of video generation models available, the produced videos often share common visual characteristics. This strongly motivates the need for an AIGC video evaluation method that achieves sufficient generalization capability with minimal training data. \textbf{Third}, there is no effective interaction between existing visual quality and generation models, as generation and understanding are mutually decoupled. This leads to the understanding model cannot obtain dynamic enhancement during the optimization process of the generation model, nor can it achieve a balance between the generalization and targeted capability.

To achieve this demand, we resort to the Group Relative Policy Optimization (GRPO)~\cite{guo2025deepseek}. As an outcome-driven reinforcement learning method, GRPO eliminates the need for an extra critic model and explicit reasoning processes during training, reducing dependence on human-labeled data and enhancing generalization. Although widely used in various vision tasks~\cite{shen2025vlm, feng2025video, xu2025avatarshield, xu2024fakeshield}, GRPO has two key issues in AIGC video evaluation: limited multi-dimension analysis and poor temporal information handling. 
Specifically, we propose a GRPO-based AIGC video quality understanding model. Through image scoring warm-up, VLMs gain preliminary understanding of image quality. By incorporating temporal modeling rewards and task-specific rewards, VLM is encouraged to acquire general video scoring and preference comparison capabilities while capturing temporal cues. Furthermore, we conduct alternating optimization between specific video generation models and understanding models to foster mutual promotion. Fig.~\ref{teasor} shows the common application scenarios of our VQ-Insight. Our contributions are summarized as follows.

\vspace{2pt}
\noindent \ding{113}~(1) We propose VQ-Insight, a reasoning-style VLM for AIGC video quality understanding. With limited data, VQ-Insight can effectively simulate human preferences and perform multi-dimension scoring, providing effective feedback for video generation.

\vspace{2pt}
\noindent \ding{113}~(2) We propose a progressive video quality learning framework, which integrates image scoring warm-up, general task-specific temporal learning, and unitied finetuning of the video generation and understanding. It enables the model to progressively move from image quality understanding to temporal perception, ultimately enhancing preference accuracy for specific video generation models.

\vspace{2pt}
\noindent \ding{113}~(3) We design a multi-dimension scoring reward and preference comparison reward, complemented by a temporal modeling and length control reward to effectively enhance the model's capability in temporal perception.

\vspace{2pt}
\noindent \ding{113}~(4) Extensive experiments show that our approach outperforms state-of-the-art methods across AIGC video preference comparison, multi-dimension scoring and even natural video scoring. Additionally, our method can be applied to alignment and editing tasks of video generation models, achieving considerable gains.

\section{Related Works}
\subsection{VLM-based Video Quality Understanding}

VLM-based quality assessment approaches~\cite{you2024depicting, you2025teaching, wu2024q, li2025q} can combine both reasoning capabilities of large language models and its powerful score regression abilities, achieving great success. In the field of video quality assessment, VQA-Scorer~\cite{jia2024vqa} introduced the SlowFast-R50~\cite{feichtenhofer2019slowfast} encoder to enhance motion capturing capabilities, and applied instruction tuning to the MLLM to guide the model to focus more on the description of low-level visual cues. In particular, evaluating AI-generated videos generally requires more complex fine-grained analyses. For instance, VideoScore~\cite{he2024videoscore} enabled automatic video quality assessment by training a VLM on the large-scale, multi-aspect human-annotated dataset VideoFeedback. VisionReward~\cite{xu2024visionreward} employed a hierarchical visual assessment framework and multi-dimension consistent preference learning to capture fine-grained human preferences for both image and video generation. UnifiedReward~\cite{wang2025unified} introduced a unified preference learning framework to enable joint pairwise ranking and pointwise scoring for multimodal generation and understanding.

\subsection{Preference Learning in Generative Models}

Recently, an increasing number of studies~\cite{wang2024lift} have explored aligning generative models with human preferences using methods such as DPO~\cite{rafailov2023direct} and GRPO~\cite{guo2025deepseek}. For example, VideoDPO~\cite{liu2025videodpo} proposed the OmniScore video evaluation pipeline, which is used to generate win-lose pairs for subsequent direct preference optimization. VADER~\cite{prabhudesai2024video} utilized a online reward model to finetune the video generation model. VideoAlign~\cite{liu2025improving} constructed a large-scale, multi-dimension human preference annotation dataset and proposed the VideoReward model, extending existing DiffusionDPO~\cite{wallace2024diffusion} to flow-based models for more fine-grained alignment. Flow-GRPO~\cite{liu2025flow} integrated online RL method into flow matching by leveraging ODE-to-SDE conversion and a denoising reduction strategy to improve generation performance. However, these methods are still limited by the accuracy and generalization issues of the reward model, making them inaccurate in effectively assisting diffusion models to learn human preferences.

\section{Methodology}

\begin{figure*}[t!]
	\centering    \includegraphics[width=1.0\linewidth]{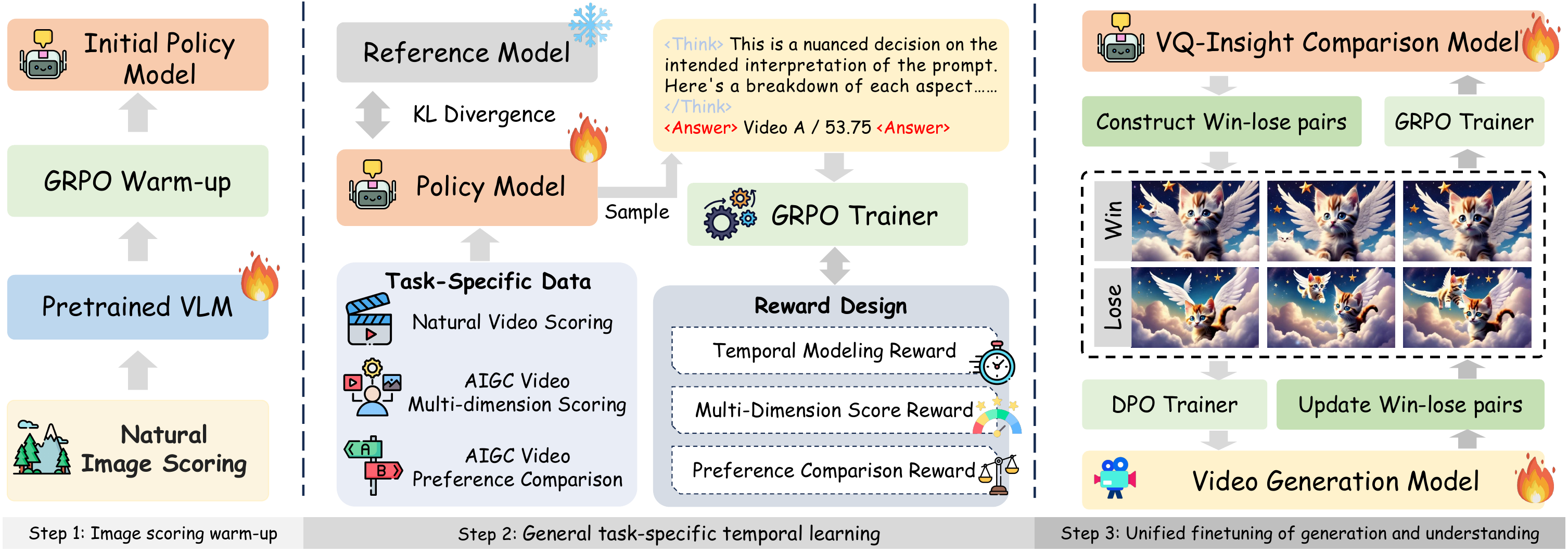}
	\vspace{-10pt}
	\caption{Illustration of the proposed VQ-Insight and our progressive visual reinforcement learning framework. In stage 1, we use the image scoring task and GRPO to warm up the pre-trained VLM; in stage 2, we employ temporal modeling rewards and task-specific rewards to enable the policy model to learn general tasks and temporal patterns; in stage 3, we jointly and alternately finetune the VQ-Insight comparison model and the video generation model, achieving a mutually beneficial effect.}
	\label{VQ-Insight}
\vspace{-10pt}
\end{figure*}

\subsection{Preliminaries}
Group Relative Policy Optimization (GRPO) is a recent advanced reinforcement learning framework for LLMs and VLMs. Distinct from proximal policy optimization like PPO which rely on a dedicated value-critic to estimate policy quality, GRPO eliminates an explicit critic by leveraging relative comparison among grouped responses. Concretely, for each query \(q\), GRPO samples a set of \(N\) candidate outputs \(\{o_1, o_2, \dots, o_N\}\) from the current or previous policy $\pi_{\theta_{\text{old}}}$. Each output receives rewards $\{r_1, r_2, \ldots, r_N\}$ based on task-specific functions, and GRPO computes the normalized advantage for each response as their reward’s deviation from the group mean, scaled by the standard deviation:
\begin{equation}
\hat{A}_{i} = \frac{r_i - \operatorname{mean}(\{r_1, r_2, \ldots, r_N\})}{\operatorname{std}(\{r_1, r_2, \ldots, r_N\})}.
\end{equation}
After obtaining the relative advantage $\hat{A}_{i}$, GRPO computes the likelihood ratio of each
response under the new policy $\pi_{\theta_{\text{new}}}$ and the old policy $\pi_{\theta_{\text{old}}}$, and clips this ratio into the interval $[1-\delta, 1+\delta]$ to prevent overly large updates and unstable training. The policy is then updated to increase the likelihood of responses with higher relative advantage, while penalizing large deviations from a given reference policy via a KL divergence term. The objective can be expressed as: 
\begin{align}
\mathcal{J}_{GRPO}(\theta) &= \mathbb{E}_{[q \sim Q, o_{i} \sim \pi_{\theta_{\text{old}}}(o|q)]}\left\{
\min \left[
    \rho_i\hat{A}_i,  \right.\right. \\ \notag
&\left.\left.\operatorname{clip}(\rho_i, 1-\delta, 1+\delta)\hat{A}_i
\right] - \beta \cdot\mathbb{D}_{\mathrm{KL}}[\pi_\theta \| \pi_{\mathrm{ref}}]
\right\},
\end{align}
where \(\rho_i = \pi_{\theta_{\text{new}}}(o_i~|~q)/{\pi_{\theta_{\text{old}}}(o_i~|~q)}\) denotes the update ratio between new and old policy for response \(o_i\), \(\delta\) controls update stability, and \(\beta\) weights the KL-regularization relative to the reference model. $Q$ denotes the question set. In essence, GRPO allows efficient and stable policy improvement by directly contrasting batches of model responses, enabling high-quality finetuning without large-scale human annotation or extra value models.

\subsection{Overview of Our VQ-Insight}

\textbf{Motivation:} Previous approaches using VLMs for video quality assessment either relied on scoring labels without reasoning processes or required explicitly constructed Chain-of-Thought (CoT) data using powerful foundational models (GPT-4o), thus consuming substantial resources. However, forceful SFT or cold-start tend to impair the general understanding capability of these models. In contrast, we hypothesize that as a heuristic and self-discovery training approach, reinforcement learning can be employed throughout the entire optimization process. Furthermore, we observe that inevitable annotator biases across diverse data sources and scoring tasks. Blindly mixing all biased data together during training can significantly harm the model's performance on each individual data domain. Thus, designing a training pipeline that progressively transfers biased knowledge from simple to complicated, general to specific scenarios emerges as the key challenge to address for training robust VLMs for video quality assessment.

We propose a curriculum-style progressive visual reinforcement learning strategy consisting of three stages: image scoring warm-up, general task-specific temporal learning, and united finetuning of generation and understanding. At each stage, we flexibly handle different tasks and data by employing tailored reward functions and training strategies, guiding the model to progressively focus on spatial relationships, temporal modeling, and text-video alignment.

\subsection{Image Scoring Warm-up}
\label{sec:imagescore}
Image quality understanding forms the foundational basis for video quality comprehension. At this stage, our main goal is to help the model learn the reasoning and response formats while improving its spatial understanding of images. Thus, as illustrated in Fig.~\ref{VQ-Insight}, we warm up a general pretrained VLM using an image scoring task to obtain the initial policy model. Specifically, we employ two distinct reward functions: a format reward and a image scoring reward. The format reward encourages the model to explicitly provide the reasoning between the \texttt{<think>} and \texttt{</think>} tags, and the numerical quality score between the \texttt{<answer>} and \texttt{</answer>} tags. Meanwhile, the image scoring reward is implemented as a continuous absolute norm ($\ell_1$-norm) to guide accurate score prediction. Given predicted score of the $i$-th response $s^{pred}_i$ and its ground truth $s^{gt}$, the reward value $r^{score}_i$ is calculated as follows.
\begin{equation}
    r^{score}_i = 1 - \|s^{pred}_i - s^{gt}\|_1.
\end{equation}

After warming up, the policy model is better able to understand image structures and visual quality, shifting its descriptive focus from high-level semantic information towards low-level details, thus facilitating subsequent task-specific optimization.

\subsection{General Task-Specific Temporal Learning}
\label{sec:temporal}
After gaining preliminary image understanding capability, we further require our VQ-Insight to move towards temporal modeling and task-specific learning. As plotted in Fig.~\ref{VQ-Insight}, in the following sections, we mainly focus on three tasks: natural video quality assessment, AIGC multi-dimension scoring, and AIGC preference comparison.  

\subsubsection{Temporal Modeling Reward:} In temporal learning, to encourage the model to assess video quality based on temporal cues, we consider using random shuffling operations to evaluate whether the model possesses sufficient temporal awareness. Specifically, given a question \( q \), we first convert the instruction into text tokens and the video into visual tokens, then concat them as input to the VLM, obtaining a set of answers \( o^{seq} \). Meanwhile, we randomly shuffle the tokens derived from video frames, feed these shuffled tokens into the policy model, and obtain another set of candidate answers \( o^{rand} \). Assuming that the model's predictions after shuffling should significantly differ from the ground truth, we compute the probability \( w^{seq} \) of giving the correct answer with sequentially ordered tokens and the probability \( w^{rand} \) with randomly shuffled tokens. If \( w^{seq}_i \) of the $i$-th response is significantly greater than \( w^{rand}_i \), we can conclude the model has successfully captured temporal information and can assign it a reward value $r^{temp}_i$ as compensation. Formally, the process is as follows.
\begin{equation}
    r^{temp}_i = \alpha~\text{ if }~w^{seq}_i > \mu \cdot w^{rand}_i,~\text{ else } 0,
\end{equation}
where $\alpha$ and $\mu$ respectively denote the hyper-parameters and set to $0.3$ and $0.8$, respectively.

\subsubsection{Length Control Reward:} To control the completion length of the policy model and avoid overthinking or underthinking, we introduce a length control reward. Specifically, if the length of the model's answer $o_i$ falls within a predefined interval $[l_{min}, l_{max}]$, we grant an additional reward $r^{len}_{i}$ for the $i$-th response.
\begin{equation}
    r^{len}_{i} = \gamma~\text{ if }~l_{min} < \operatorname{len}(o_i) < l_{max}, ~\text{ else } 0, 
\end{equation}
where $\gamma$ is set to $0.1$. $l_{min}$ and
 $l_{max}$ are empirically set to $320$ and $512$. We observe that introducing the length control reward lead to an ``aha moment" in the large model's understanding of video quality, and the model paid greater attention to temporal modeling during reasoning.

\subsubsection{Multi-Dimension Scoring Reward:} Unlike scoring images, video quality assessment often requires consideration from multiple aspects. Following UGVQ~\cite{zhang2024benchmarking}, we mainly focus on three aspects: spatial quality, temporal quality, and text-video alignment, each represented by a Mean Opinion Score (MOS). Given a query \( q \), as shown in Fig.~\ref{VQ-Insight}, we prompt the VLM to directly output a set of scores with \( M \) dimensions $\{v^{pred}_{i,j}\}^{M}_{j=1}$ in the $i$-th response. Similar to image scoring warm-up stage, we also adopt the $\ell_1$ norm to fit scores in each dimension separately.
\begin{equation}
    r^{multd}_i = 1 - \sum^M_{j=1}\lambda_j \|v^{pred}_{i,j} - v^{gt}_j\|_1,
\label{multidimension}
\end{equation}
where $v^{gt}_j$ denotes the ground truth score of the $j$-th dimension. $\lambda_j$ is used to balance the weights of different dimensions. Note that, for natural video quality assessment, we still adopt a single-dimension reward ($M$~=~1) due to the limitations of existing datasets.

\subsubsection{Preference Comparison Reward:} Rather than obtaining an absolute score, it is often more meaningful to directly provide the relative ranking between two generated results. To this end, we introduce the preference comparison reward. Specifically, given two input videos, we first convert them into visual tokens separately and feed them into the VLM to produce a group of preference choice $c^{pred}_i$. If the chosen answer $c^{pred}_i$ matches the ground truth $c^{gt}$, we set \( r^{comp}_i \) to $1$. To enhance the model's multimodal scoring capability while preserving its general understanding ability, we include additional AIGC visual question answering data as an auxiliary task. Given an input video, the VLM is sequentially asked questions such as ``Is the motion pattern in this video reasonable?" The large model is required only to answer ``yes" or ``no". If the model answers correctly, we assign a reward value \( r^{comp}_i \) of $1$. The rewards for these two tasks can be expressed uniformly as follows.

\begin{equation}
   r^{comp}_i = 1~\text{ if }~c^{pred}_i = c^{gt},~\text{ else } 0,   
\end{equation}
where \( c^{gt}_i \) is ``video A" or ``video B" in the preference comparison task, and ``yes" or ``no" in the VQA task. Ultimately, we combine the temporal-aware rewards and task-specific rule-based rewards to jointly optimize our VQ-Insight with robust and general video understanding capabilities.

\subsection{United Finetuning of Generation and Understanding}
\label{sec:united}
\textbf{Motivation:} To apply VQ-Insight to the video generation model, a common practice is to employ Direct Preference Optimization (DPO) to align generated outputs with human preferences. However, since DPO is an offline RL method, its preference dataset cannot dynamically evolve alongside the generation model during optimization, causing the generation model to quickly reach a performance ceiling. On the other hand, video generation models finetuned by DPO typically possess stronger generative capabilities, further widening the gap between the newly generated positive samples and the original samples, thus enabling updates to our preference dataset. This updated preference dataset has the potential to enhance VQ-Insight's understanding capabilities, allowing it to better focus on preference comparisons specific to certain generative models.

Specifically, we first use the video generation model $\mathcal{G}_{\theta}$ to produce \( N \) candidate videos \(\mathcal{X} = \{\boldsymbol{x}_1, \boldsymbol{x}_2, \dots, \boldsymbol{x}_N\}\). Then, we form pairs of these videos and employ the VQ-Insight (stage 2) $\mathcal{D}_{\theta}$ to conduct \( \binom{N}{2} \) preference estimations. Finally, by counting the number of times each candidate is preferred, we identify the most- and least-chosen videos within the set, thus obtaining win-lose pairs $\mathcal{C}=\{(\boldsymbol{x}^{w}, \boldsymbol{x}^{l})_k\}_{k=1}^K$.
\begin{equation}
    \boldsymbol{x}^{w}=\operatorname*{argmax}_{\boldsymbol{x}_i,\boldsymbol{x}_j~\in~\mathcal{X}}\mathcal{D}_{\theta}(\boldsymbol{x}_i, \boldsymbol{x}_j),~\boldsymbol{x}^{l} = \operatorname*{argmin}_{\boldsymbol{x}_i,\boldsymbol{x}_j~\in~\mathcal{X}}\mathcal{D}_{\theta}(\boldsymbol{x}_i, \boldsymbol{x}_j).
\end{equation}
Following the approach of DiffusionDPO~\cite{wallace2024diffusion}, we can optimize \( \mathcal{G}_{\theta} \) by comparing the noise prediction differences between the finetuned model and the reference model. The loss function is defined as follows.
\begin{multline}
    \ell_{dpo}
    = - \mathbb{E}_{
    (\boldsymbol{x}^w, \boldsymbol{x}^l)\in\mathcal{C},\boldsymbol{x}^w_{t},\boldsymbol{x}^l_{t}, t\sim \mathcal{U}(0,T)
    }
    \log\sigma \left(-\Gamma(\lambda_t) \left(\right. \right.\\
    \| \boldsymbol{\epsilon}^w -\mathcal{G}_\theta(\boldsymbol{x}_{t}^w,t)\|^2_2 - \|\boldsymbol{\epsilon}^w - \mathcal{G}_{ref}(\boldsymbol{x}_{t}^w,t)\|^2_2 \\
    \left. \left.  - \left( \| \boldsymbol{\epsilon}^l -\mathcal{G}_{\theta}(\boldsymbol{x}_{t}^l,t)\|^2_2 - \|\boldsymbol{\epsilon}^l - \mathcal{G}_{ref}(\boldsymbol{x}_{t}^l,t)\|^2_2\right)
    \right)\right),
    \label{dpo}
\end{multline}
where \(\boldsymbol{x}^*_t\) is the noised latents obtained by adding noise to \( \boldsymbol{x}^* \) at timestep \( t \). $\boldsymbol{\epsilon}^{*}$ is the preset noise sampled from the distribution $q(\boldsymbol{x}^{*}_t|\boldsymbol{x}^{*}_0)$. $\Gamma(\cdot)$ and $\lambda_t$ respectively denote a weighting function and the SNR ratio. $\sigma$ is the logistic function. After obtaining the finetuned video generation model \(\mathcal{G}^{\prime}_{\theta}\) according to Eq.~\ref{dpo}, we generate \(N\) additional samples via this updated model. Then, we utilize VQ-Insight $\mathcal{D}_{\theta}$ to select the best candidate $\hat{\boldsymbol{x}}^w$ from these newly generated samples, pairing it with the previous lose sample $\boldsymbol{x}^l$ to form a new preference set $\hat{\mathcal{C}}=\{(\hat{\boldsymbol{x}}^w,\boldsymbol{x}^l)_m\}_{m=1}^M$. As shown in Fig.~\ref{VQ-Insight}, we combine \(\hat{\mathcal{C}}\) with the original data and continue to finetune \(\mathcal{D}_{\theta}\) via the training strategy in Stage 2, resulting in a preference model \(\mathcal{D}^{\prime}_{\theta}\) specialized for \(\mathcal{G}_\theta\). Finally, the updated model $\mathcal{D}^{\prime}_{\theta}$ is used to generate a new preference set for video DPO, resulting in a better generation model \(\mathcal{G}^{\prime\prime}_{\theta}\).

\section{Experimental Results}

\subsection{Experimental Setup}
\subsubsection{Dataset and Metrics:} We use 7k images in KonIQ~\cite{hosu2020koniq} to perform warm-up. In Stage 2, only 2k comparison videos~\cite{xu2024visionreward} and 1k VQA data~\cite{xu2024visionreward} are employed to train for the preference comparison task. LGVQ~\cite{zhang2024benchmarking} and LSVQ~\cite{ying2021patch} are used for the AIGC multi-dimension scoring and natural video scoring tasks. In Stage 3, we choose T2V-Turbo~\cite{li2024t2v} as generation models and select 5k prompts from Vidprom~\cite{wang2024vidprom} for united finetuning. To evaluate the preference comparison capability of our model, Gen-AI~\cite{jiang2024genai} and MonetBench~\cite{xu2024visionreward} are employed, with preference selection accuracy used as the evaluation metric. Moreover, LGVQ is utilized to assess our model's performance in multi-dimension scoring. LSVQ-Test, LSVQ-1080p, LIVE-VQC~\cite{sinno2018large}, and KonViD-1k~\cite{hosu2017konstanz} are adopted to evaluate the model's natural video quality scoring ability, using Pearson Linear Correlation Coefficient (PLCC), Spearman Rank-order Correlation Coefficient (SRCC), and Kendall Rank-order Correlation Coefficient (KRCC) as the evaluation metrics. For video generation tasks, we employ VBench~\cite{huang2024vbench} to evaluate the generation quality of the finetuned models.

\begin{figure*}[t!]
	\centering    \includegraphics[width=1.0\linewidth]{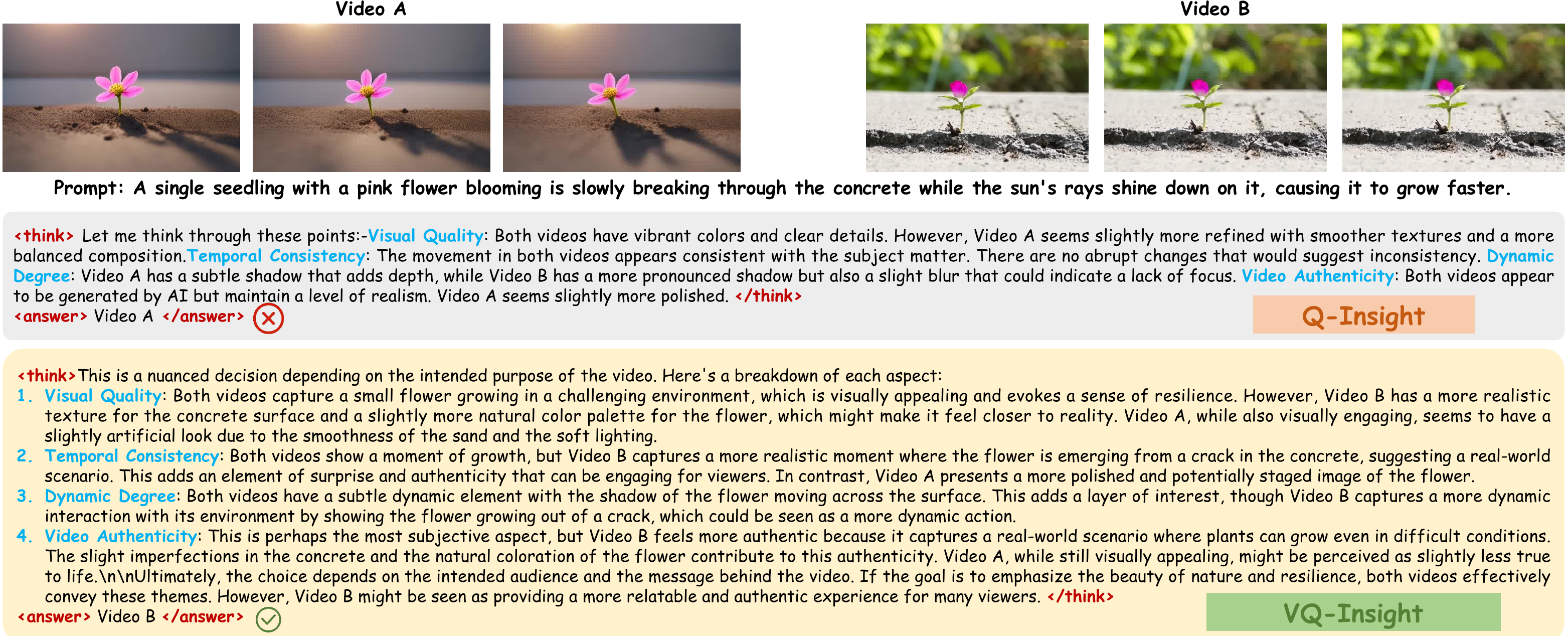}
	\vspace{-15pt}
	\caption{Reasoning process comparison between Q-Insight and the proposed VQ-Insight on preference comparison tasks.}
	\label{comparison_vqinsight}
    \vspace{-10pt}
\end{figure*}

\subsubsection{Implementation Details:}Qwen-2.5-VL-7B-Instruct~\cite{bai2025qwen2} is used as our pretrained VLM. The generation number $N$ and the wieght of KL penalty $\beta$ in the GRPO trainer are $8$ and $0.001$. The model is trained for $3$ epochs on $8$ NVIDIA A100 80G GPUs, with a learning rate of $1$$\times$$10^{-6}$. $\lambda_j$ in Eq.~\ref{multidimension} is set to $1$.

\subsection{AIGC Video Preference Comparison}

\begin{table}[t!]
\centering
\renewcommand{\arraystretch}{1.3}
\caption{Preference comparison between our VQ-Insight and other competitive methods across GenAI and MonetBench dataset evaluated using tau and diff scores.}
\vspace{-5pt}
\resizebox{1.\linewidth}{!}{
\begin{NiceTabular}{c|cccc}
\CodeBefore
\tikz \fill [gray!10] (1-|1) rectangle (2-|6);   
\tikz \fill [gray!10] (2-|1) rectangle (3-|6); 
\Body

\shline
\multirow{2}{*}{Dataset} & \multicolumn{2}{c}{GenAI} & \multicolumn{2}{c}{MonetBench} \\ \cline{2-5} 
& tau & diff & tau & diff \\ \hline
VQAScore~\cite{lin2024evaluating} & 46.96 & 69.14 & 54.00 & 59.39 \\
VideoScore~\cite{he2024videoscore} & 47.43 & 70.50 & 49.10 & 54.90 \\
VisionReward~\cite{xu2024visionreward} & 46.68 & 68.86 & \secondbest{59.40} & \secondbest{72.44} \\
VideoReward~\cite{liu2025improving} & 45.84 & 69.00 & 53.60 & 59.88 \\
Qwen-SFT~\cite{bai2025qwen2} & 40.69 & 59.43 & 59.20 & 72.07 \\
Q-Insight~\cite{li2025q} & 47.52 & 70.43 & 49.60 & 60.37 \\
UnifiedReward~\cite{wang2025unifiedcot} & \secondbest{49.67} & \secondbest{74.42} & 52.10 & 62.56 \\
VQ-Insight & \best{50.80} & \best{75.71} & \best{61.20} & \best{74.51} \\ \shline
\end{NiceTabular}}
\label{table:comparison}
\vspace{-10pt}
\end{table}

To evaluate the performance of our VQ-Insight in preference comparison, we selected classic methods such as VQAScore~\cite{jia2024vqa}, SFT-based VLM methods including VideoScore~\cite{he2024videoscore}, VisionReward~\cite{xu2024visionreward}, VideoReward~\cite{liu2025improving}, and Qwen-SFT~\cite{bai2025qwen2}, as well as RL-based VLM methods like Q-Insight~\cite{li2025q} and UnifiedReward-Think~\cite{wang2025unifiedcot}. For a fair comparison, we use the publicly available pretrained weights and the same evaluation scripts for testing.

As reported on Tab.~\ref{table:comparison}, our method surpasses existing SOTA approaches such as VisionReward~\cite{xu2024visionreward} on both ``tau'' and ``diff'' condition for computing preference accuracy, which demonstrates the effectiveness and strong generalization capability of our method. Note that ``tau'' uses a tau-corrected result~\cite{deutsch2023ties} for preference accuracy, while ``diff'' excludes Tie cases. Fig.~\ref{comparison_vqinsight} further demonstrates the superiority of our method. Compared to the recent reasoning-based image scoring model Q-Insight, VQ-Insight can provide detailed and accurate explanations and analyses from four perspectives: visual quality, temporal consistency, dynamic degree, and video authenticity, while delivering more accurate preference choice results. This can be attributed to our refined reward design and progressive training paradigm.

\subsection{AIGC Video Multi-Dimension Scoring}
\begin{table*}[t!]
\centering
\renewcommand{\arraystretch}{1.3}
\caption{SRCC, KRCC, PLCC Comparison between VQ-Insight and other competitive methods across spatial quality, temporal quality, and text-video alignment dimensions.}
\vspace{-5pt}
\resizebox{1.\linewidth}{!}{
\begin{NiceTabular}{c|c|c|c|c|c|c|c|c|c}
\CodeBefore
\tikz \fill [gray!10] (1-|1) rectangle (2-|11);   
\tikz \fill [gray!10] (2-|1) rectangle (3-|11); 
\Body

\shline
\multirow{2}{*}{Method} & \multicolumn{3}{c}{Spatial Quality} & \multicolumn{3}{c}{Temporal Quality} & \multicolumn{3}{c}{Text-Video Alignment} \\ \cline{2-10}
& SRCC & KRCC & PLCC & SRCC & KRCC & PLCC & SRCC & KRCC & PLCC         \\ \hline
CLIP-IQA~\cite{wang2023exploring} & 0.684 & 0.502 & 0.709 & - & - & - & -& - & - \\ 
FastVQA~\cite{wu2022fast} & - & - & - & 0.849 & 0.672 & 0.878 & -& - & -      \\ 
CLIPScore~\cite{hessel2021clipscore} & - & - & - & - & - & - & 0.446 & 0.301 & 0.453  \\
UGVQ~\cite{zhang2024benchmarking} & \secondbest{0.764} & \secondbest{0.571} & \secondbest{0.793} & \secondbest{0.894} & \secondbest{0.703} & \secondbest{0.910}  & 0.545 & 0.391 & 0.569        \\
UnifiedReward~\cite{wang2025unifiedcot} & 0.580 &0.432 & 0.594 & 0.466 &0.330 & 0.500 & 0.589 &0.433 & 0.589 \\
Qwen-SFT~\cite{bai2025qwen2}  & 0.687 &0.520 & 0.735 & 0.723 & 0.539 & 0.750 & \secondbest{0.605} &\secondbest{0.462} & \secondbest{0.660} \\
VQ-Insight (Ours)  & \best{0.823} &\best{0.640} & \best{0.844} & \best{0.911} & \best{0.744} & \best{0.927} & \best{0.825} &\best{0.652} & \best{0.836} \\ \shline
\end{NiceTabular}}
\vspace{-5pt}
\label{table:score}
\end{table*}

To evaluate the performance of our method on fine-grained video quality assessment, we conduct training and testing on the LGVQ dataset. Following the setup of LGVQ~\cite{zhang2024benchmarking}, we mainly consider three dimensions: spatial quality, temporal quality, and text-video alignment. For comparison, we select several metrics that are limited to a single dimension, such as CLIP-IQA~\cite{wang2023exploring}, CLIP-Score~\cite{hessel2021clipscore}, and FAST-VQA~\cite{wu2022fast}. In addition, we include more comprehensive scorers such as UGVQ~\cite{zhang2024benchmarking}, UnifiedReward-Think~\cite{wang2025unifiedcot}, and Qwen-SFT~\cite{bai2025qwen2}.

As reported on Tab.~\ref{table:score}, our method significantly outperforms the SOTA approaches UGVQ and Qwen-SFT across all dimensions, achieving a well-balanced performance in fine-grained video assessment. On the spatial quality dimension, our method surpasses UGVQ by 0.051 and 0.059 on PLCC and SRCC, respectively. Furthermore, on the text-video alignment dimension, our approach achieves an improvement of up to 0.2 over previous methods, demonstrating that our progressive reinforcement learning strategy effectively preserves the VLM's general language understanding ability and world knowledge priors. Fig.~\ref{teasor} presents an example of VQ-Insight performing fine-grained scoring. Our VQ-Insight can comprehensively consider the spatial and temporal quality of the video while analyzing alignment based on the given prompt, ``Students memorize lessons in the classroom.''

\subsection{Natural Video Scoring}
In addition to AIGC video evaluation, our method can also be extended to natural video scoring. In the stage~1, we warm up the model with a natural image quality assessment task, and in stage~2, we train our VQ-Insight on the LSVQ dataset~\cite{ying2021patch}. We conduct experiments on four datasets, namely LSVQ-Test, LSVQ-1080p, Live-VQC, and Konvid-1k~\cite{hosu2017konstanz}. The comparison baselines include the classic video quality assessment model Fast-VQA~\cite{wu2022fast}, as well as VLM-based methods such as Q-Align~\cite{wu2024q}, Q-Instruct~\cite{wu2024qinstruct}, VQA$^2$~\cite{jia2024vqa}, and Q-Insight~\cite{li2025q}.
\begin{table}[t!]
    \centering
    \renewcommand{\arraystretch}{1.3}
    \caption{PLCC and SRCC comparisons on the natural video scoring tasks between our VQ-Insight and other methods.}
    \vspace{-5pt}
    \resizebox{1.\linewidth}{!}{
    \begin{NiceTabular}{c|c|cccc}
\CodeBefore
\tikz \fill [gray!10] (1-|1) rectangle (2-|7);    
\Body
        \shline
        Model & Metric & \begin{tabular}[c]{@{}c@{}}LSVQ\\-Test\end{tabular} & \begin{tabular}[c]{@{}c@{}}LSVQ\\-1080p\end{tabular} &\begin{tabular}[c]{@{}c@{}}LIVE\\-VQC\end{tabular} & \begin{tabular}[c]{@{}c@{}}KonViD\\-1k\end{tabular} \\
        \midrule
        Fast-VQA  &PLCC & 0.878 & 0.810 & 0.815 & 0.857              \\
        ~\cite{wu2022fast} &SRCC & 0.874  & 0.765 & 0.769              & 0.859              \\
        Minimalist-VQA &PLCC &0.872 &0.818 &0.812 &0.861 \\
        ~\cite{sun2024analysis} &SRCC &0.880 &0.769 &0.765 &0.859 \\
        mPLUG-owl-2  &PLCC &0.434   &0.422  &0.459  &0.532  \\
        ~\cite{ye2024mplug} &SRCC &0.422  &0.398  &0.450  &0.532  \\
        Q-Align &PLCC & \best{0.882}  & \best{0.833}  & 0.813  & \secondbest{0.876}              \\
        ~\cite{wu2024q} &SRCC & \best{0.883}  & \secondbest{0.758}  & 0.777  & \secondbest{0.865}              \\
        Q-Instruct &PLCC & 0.580  &0.640   & 0.673  & 0.520              \\
        ~\cite{wu2024qinstruct} &SRCC &0.602  &0.644 &0.660   & 0.492              \\
        VQA$^2$ &PLCC & 0.856    & 0.819         & \secondbest{0.823}         & 0.844         \\
        ~\cite{jia2024vqa} &SRCC & \secondbest{0.882}         & 0.760         & \secondbest{0.776} & 0.833         \\
        Q-Insight  &PLCC & 0.639 & 0.648   &  0.708 & 0.753      \\
        ~\cite{li2025q} &SRCC & 0.644 & 0.601   &  0.624 & 0.751      \\
        VQ-Insight &PLCC  & \secondbest{0.876}  & \secondbest{0.823}  & \best{0.835} & \best{0.884} \\
        (Ours)  &SRCC & 0.875  & \best{0.786}  & \best{0.790} & \best{0.875}  \\
        \shline
    \end{NiceTabular}}
    \label{tab:performance_comparison}
    \vspace{-10pt}
\end{table}
As reported on Tab.~\ref{tab:performance_comparison}, our VQ-Insight achieves the best or near-best PLCC and SRCC on the LIVE-VQC, KonViD-1k, and LSVQ-1080p datasets, demonstrating its strong generalization ability on out-of-domain data. On the in-domain dataset LSVQ-Test, the performance of VQ-Insight is comparable to that of the SOTA methods Q-Align and VQA\textsuperscript{2}. This can be attributed to our progressive visual reinforcement learning strategy and temporal modeling reward used by our VQ-Insight, which better capture temporal cues and unlock the model's potential for quality understanding. Furthermore, as shown in Fig.~\ref{teasor}, the results produced by VQ-Insight effectively describe the content of the video and identify the imbalance between the theme and the environment.

\begin{table}[t!]
    \centering
    \renewcommand{\arraystretch}{1.2}
    \caption{Ablation study on different components of our VQ-Insight for the AIGC multi-dimension scoring tasks.}
    \vspace{-5pt}
    \resizebox{1.0\linewidth}{!}{
    \begin{NiceTabular}{cccc|ccc}
    \CodeBefore
\tikz \fill [gray!10] (1-|1) rectangle (2-|8);    
\Body
        \shline
        Case & 
        Warm-up & 
        TMR & 
        LCR & 
        PLCC & 
        KRCC &
        SRCC \\
        \hline
        (a) & \ding{55} & \ding{51} & \ding{51} & 0.716 &0.518 & 0.690 \\
        (b) & \ding{51} & \ding{55} & \ding{51} & 0.787 &0.590 & 0.761 \\
        (c) & \ding{51} & \ding{51} & \ding{55} & 0.819 & 0.614 & 0.791 \\
        (d) & \ding{51} & \ding{51} & \ding{51} & 0.869 &0.679 & 0.853 \\
        \shline
    \end{NiceTabular}}

    \label{tab:ablation_study}
\end{table}

\begin{table}[t!]
\centering
\renewcommand{\arraystretch}{1.2}
\caption{Ablation study on different components of our VQ-Insight for the AIGC preference comparison tasks.}
\vspace{-5pt}
\resizebox{1.0\linewidth}{!}{
\begin{NiceTabular}{ccc|cc|cc}
\CodeBefore
\tikz \fill [gray!10] (1-|1) rectangle (2-|8); \tikz \fill [gray!10] (2-|1) rectangle (3-|8); 
\Body 
\shline
\multirow{2}{*}{Case} & \multirow{2}{*}{LCR} & \multirow{2}{*}{UF} & \multicolumn{2}{c|}{GenAI} & \multicolumn{2}{c}{MonetBench} \\ \cline{4-7} 
& & & tau & diff & tau & diff \\
\hline
(e) & \ding{56} &\ding{52} & 45.74 & 68.14  & 60.00 & 73.05           \\ 
(f) & \ding{52} &\ding{56} & 50.14 & 75.14  & 60.20 & 73.29           \\
(g) & \ding{52} &\ding{52} & 50.80 & 75.71  & 61.20 & 74.51           \\ \shline
\end{NiceTabular}}
\label{tab:aba_comp}
\vspace{-10pt}
\end{table}

\begin{figure*}[t!]
	\centering    \includegraphics[width=1.0\linewidth]{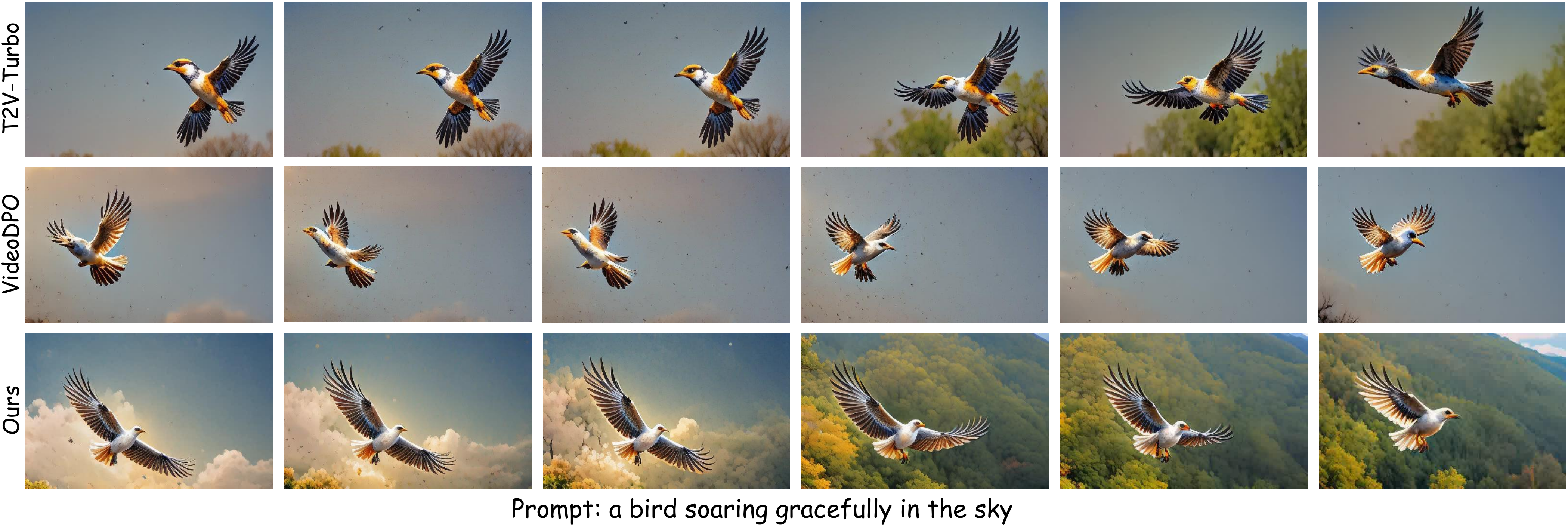}
	\vspace{-15pt}
	\caption{Generation result comparisons between our method and other competitive methods. The video generation model finetuned with VQ-Insight can mitigate the issue of birds with multiple wings, while also producing more vibrant colors and increased dynamic degrees.}
	\label{Videogeneration}
    \vspace{-10pt}
\end{figure*}

\subsection{Ablation Studies}
To validate the contributions of each component in our VQ-Insight, we design some variations and retrain them for the tasks of AIGC multi-dimension scoring and preference comparison by targeting image scoring warm-up (Warm-up), temporal modeling reward (TMR), length control reward (LCR), and unified finetuning (UF).

Tab.~\ref{tab:ablation_study} and Tab.~\ref{tab:aba_comp} report our results. We observe that skipping the image scoring warm-up step and directly starting training from the pre-trained weights of Qwen2.5-VL results in noticeable degradation in both scoring and comparison performance. This highlights the importance of the warm-up phase for establishing the model's perception of visual quality. Removing the TMR leads to a significant drop of 0.082 in PLCC for VQ-Insight's performance on multi-dimension scoring tasks, demonstrating the critical role of TMR in helping the model capture motion patterns. Since performing temporal shuffle on two videos simultaneously can lead to preference label confusion, we do not validate the effectiveness of TWR on the preference comparison task. Additionally, we find that the length control reward (LCR) can better guide the model to produce reasonably detailed reasoning results while outputting more accurate scores or preference choices. Finally, when removing the unified finetuning strategy for generation and understanding (case (f)), we observe that VQ-Insight's accuracy on GenAI and MonetBench datasets decreased by 0.57 and 1.22, respectively. Moreover, as shown in Tab.~\ref{tab:generation}, the performance of video generation models also experience certain degradation, which is caused by the inaccuracy of the comparison model.
\begin{table}[t!]
    \centering
    \renewcommand{\arraystretch}{1.3}
    \caption{VBench Score Comparison between our method, VideoDPO and the original T2V-Turbo.}
    \vspace{-5pt}
    \resizebox{1.\linewidth}{!}{
    \begin{NiceTabular}{c|ccc}
\CodeBefore
\tikz \fill [gray!10] (1-|1) rectangle (2-|5);   
\Body
        \shline
        Method & Overall Score & Quality Score & Semantic Score \\
        \midrule
        T2V-Turbo & 0.8095 & 0.8271 & 0.7393 \\
        VideoDPO & \secondbest{0.8167} & \secondbest{0.8367} & 0.7365 \\
        Ours-w/o UF & 0.8149 & 0.8325 & \secondbest{0.7444} \\
        Ours  & \best{0.8185} & \best{0.8368} & \best{0.7450} \\
        \shline
    \end{NiceTabular}}
    \vspace{-10pt}
    \label{tab:generation}
\end{table}

\subsection{Applications}
To validate that our VQ-Insight can effectively support generation tasks, we conduct DPO post-training on the video generation model T2V-Turbo~\cite{li2024t2v}. We select 5k prompts from Vidprom~\cite{wang2024vidprom} and use the T2V-Turbo to produce 10 results for each prompt. Subsequently, VQ-Insight is used to select the best and worst results for DPO training. The performance of our finetuned model evaluated on VBench~\cite{huang2024vbench} is shown in Tab.~\ref{tab:generation}. For VideoDPO, we use its provided preference dataset and code for re-training. It can be observed that, compared to VideoDPO and baseline results, our method achieves significant improvements in overall score and quality score. Specifically, in the semantic score metric, VQ-Insight demonstrates strong general understanding capabilities, making it particularly effective in handling this type of preference choice. As a result, our finetuned generation model shows a 0.0057 improvement over the baseline on the semantic score. Furthermore, Fig.~\ref{Videogeneration} demonstrates that our method achieves noticeable improvements over both the baseline and VideoDPO in terms of the dynamic degree, subject consistency, background richness, and color vividness.

\section{Conclusion}

In this paper, we propose VQ-Insight, a novel reasoning-style vision-language model framework for AIGC video quality assessment. By introducing a progressive learning scheme that combines image warm-up, temporal learning, and joint optimization with generation models, as well as task-specific rewards, our method achieves superior accuracy and generalization with limited data. Extensive experiments demonstrate that VQ-Insight consistently outperforms state-of-the-art baselines across multiple video scoring and comparison benchmarks and can be effectively applied to both generation alignment and content editing. Looking ahead, our unified approach sets the stage for more dynamic, human-aligned video evaluation and optimization, highlighting the potential for further integration of reinforcement learning and multimodal reasoning in the field.

\subsection{Limitations} Although our VQ-Insight can align the performance of video generation models to human preferences, the overall improvement also depends on the capability of the baseline generative model itself. We believe that our approach will continue to improve as the performance of generation models advances. Meanwhile, current methods generally use roughly the same completion length for all cases. In the future, it may be necessary to introduce more flexible length control strategies, allowing the inference length to adapt dynamically based on the difficulty of the video.

\clearpage
\bibliography{aaai2026}

\end{document}